\documentclass[conference]{IEEEtran}
\IEEEoverridecommandlockouts
\usepackage{cite}
\usepackage{amsmath,amssymb,amsfonts}
\usepackage{graphicx}
\usepackage{bm}
\usepackage{textcomp}
\usepackage{xcolor}
\usepackage{tikz}
\usepackage{pgfplots}
\def\BibTeX{{\rm B\kern-.05em{\sc i\kern-.025em b}\kern-.08em
    T\kern-.1667em\lower.7ex\hbox{E}\kern-.125emX}}
    \pgfplotsset{compat=1.18}
\begin{document}

\title{Resilient Sparse Array Radar with the Aid of Deep Learning\\

\thanks{This work is funded by the German Federal Ministry of Education and
Research (BMBF) in the course of the 6GEM Research Hub under
Grant 16KISK037.}
}

\author{\IEEEauthorblockN{Aya Mostafa Ahmed$^{\star}$~Udaya S.K.P. Miriya Thanthrige$^{\dagger}$,~Aydin Sezgin$^{\star}$ and Fulvio Gini$^{\ddagger}$} 
	\IEEEauthorblockA{\textit{$^{\star}$ Ruhr-University Bochum, Bochum, Germany.} \\
		\textit{$^{\dagger}$ University of Moratuwa, Sri Lanka.}	\\
	\textit{$^{\ddagger}$
		University of Pisa,Pisa, Italy.}\\
		\small{ aya.mostafaibrahimahmad@rub.de,~sampathk@uom.lk,~aydin.sezgin@rub.de,~fulvio.gini@unipi.it.}}
}

\IEEEpubid{ Accepted to be published in VTC2023-Spring, 2023.  \copyright\ 2023 IEEE. Personal use is permitted, but republication/redistribution requires IEEE permission.}
\maketitle

\begin{abstract}
In this paper, we address the problem of direction of arrival (DOA)
estimation for multiple targets in the presence of sensor
failures in a sparse array. Generally, sparse arrays are known
with very high-resolution capabilities, where $N$ physical sensors can resolve up to $\mathcal{O}(N^2)$ uncorrelated sources. However, among
the many configurations introduced in the literature, the arrays
that provide the largest hole-free co-array are
the most susceptible to sensor failures. We propose here
two machine learning (ML) methods to mitigate the effect of
sensor failures and maintain the DOA estimation performance and
resolution. The first method enhances the conventional spatial
smoothing using deep neural network (DNN), while the second one
is an end-to-end data-driven method. Numerical results show
that both approaches can significantly improve the performance
of MRA with two failed sensors. The data-driven method can maintain the
performance of the array with no failures at high signal-to-noise ratio (SNR).
Moreover, both approaches can even perform better than the original array
at low SNR thanks to the denoising effect of the proposed DNN.
\end{abstract}


\section{Introduction}
Resilience is a critical key performance indicator (KPI) in the upcoming 6G networks. Thus, in contrast to previous generations, the network is supposed to provide instant recovery from a disaster or failure. Within this context, its scope extends beyond the traditional approaches of robustness and reliability \cite{6G_JCS,6G}. This adds a necessary functionality which is reconfigurablity of the system to adapt to the erroneous events and automatically recover its required function in a timely manner without any human interaction \cite{robertspaper}.  Within this framework, it is most vital for 6G networks, especially the ones providing critical sensing services, to ensure resilience in connectivity, where the system is able to tolerate connectivity disruptions and failure among its components while maintaining the system's functionality.  Among these disruptions, failure in the elements of the antenna arrays is considered a challenge in resilient networks. This paper considers the aspect of resilience with respect
to direction of arrival (DOA) estimation using an antenna array with failed elements.\\ 
A very important application of antenna arrays is DOA estimation, where the spatial spectra of the received electromagnetic waves determined, to estimate the location of possible targets within the array's field of view. Optimizing the antenna locations in an array can offer high degrees of freedom (DoF) through extending a co-array aperture \cite{sparsearray}. This is due to the fact that the augmented covariance matrix of a difference co-array explores the unique correlation lags of a certain array configuration, leading to resolving more paths than the actual number of physical antennas. Sparse arrays of $N$ physical antennas can resolve up to $\mathcal{O}(N^2)$ uncorrelated paths, unlike uniform linear arrays (ULA) with the same number of antennas. 
Several array configurations were proposed in the literature, such as minimum redundancy arrays (MRA) \cite{MRA}, nested arrays \cite{nestedarrays}, and coprime arrays \cite{sparsearray}. Among the previously mentioned configurations, MRA is known to have the best estimation performance, since it is characterized by the largest hole-free difference co-array \cite{MRAoptimization}. 
Despite the high DoF offered by the sparse array, such attractive property remains susceptible to sensor failures in the array \cite{MRAoptimization}. The same coarray structure that offers the $\mathcal{O}(N^2)$ resolution property, will be partially destroyed when one or more antennas fail, leading to a large loss in the DoF compared to the ULA of similar physical antennas. This in turn causes performance degradation and possible breakdown of the overall system functionality. The authors in \cite{essentialarray} compared the robustness of different coarrays configuration and structures, using the \textit{essentialness property}, where an antenna element is said to be essential, if its deletion affects the corresponding difference coarray structure. Furthermore, the authors deduced that sparse arrays offering large difference coarray, like MRA for instance, are less robust to sensor failures since most physical antennas in the array are considered essential elements.
Towards this purpose, this paper uses sensor failures in sparse array radar as a showcase for the criticality of a resilient radar system.\\ \IEEEpubidadjcol
In general, many previous works have addressed the problem of missing data within DOA estimation. In \cite{Stoica_missingdata}, the authors proposed a maximum likelihood estimator based on Cholesky parametrization when some sensors fail to work before the measurement is complete. However, they used ULA in their model requiring a specific sequential failure pattern, which might not be practical. In \cite{SOA_1}, the authors addressed the same problem using ULA as well, however they categorized the failures into nonessential and essential failures in the antenna array. 
The former problem was solved by modeling the ULA with failed sensors as a sparse array, hence they used the remaining sensors to form the difference coarray. In the latter problem, the constructed difference coarray has some holes, in this case, they exploited the low-rank property of the signal subspace, and through trace norm minimization, they were able to recover those missing holes. However, the complexity of their approach was in terms of $\mathcal{O}(N^3)$, hence when the number of sensors is large, the algorithm will be computationally expensive to run in real-time, which would contradict the resilience requirements in 6G systems. In \cite{missingelement}, the authors addressed this problem by devising an algorithm, that assumes that all sensors are functioning in the first sampling period, then they use the devised covariance matrix of the complete array to calculate the covariance matrix of the incomplete one. This approach would be considered very beneficial in static environments. However, in a rapidly changing environment, as expected in 6G networks, the information of the complete measurements can be totally independent of the incomplete measurements in later periods, leading to false estimations. \\
In this work, we address all these aspects, to fulfill the resilience framework defined earlier. We use 
a machine learning (ML) approach to mitigate the effect of sensor failures in a sparse array. We specifically address the worst-case scenario as defined in \cite{essentialarray}, in which failures can occur in more than one essential sensor in the MRA configuration of a large aperture. As a result, the constructed coarray would consist of very small continuous ULAs, leading to degraded performance. Thus, we use ML to introduce a resilient array that is tolerant to failures in any sensor.\\
In this paper, the following notation is used. For a given matrix $\bm{A}$, we denote its transpose, Hermitian transpose, and conjugate by $\bm{A}^T$, $\bm{A}^H$, and $\bm{A}^*$, respectively. $\text{vec}(\cdot)$~represents the vectorization operation, where, it stacks the columns of a matrix to convert a matrix to a vector. Further, to represent real and imaginary components of a complex number, we use $\Re (\cdot)$ and $\Im (\cdot)$, respectively. The expected value is denoted by $\mathbb{E}(\cdot)$. Note that, we denote the Kronecker product and the Khatri-Rao product by $\otimes$ and $\bigodot$, respectively. An identity matrix with a dimension of $M \times M$ is denoted by $\bm{{I}}_M$.
\section{System Model}
    
In this work, we consider a minimum redundancy array (MRA) which contains $M$ number of sensors. We have chosen MRA because it is considered one of the most fragile configurations to failure. This is due to the fact that the MRA has the largest hole-free difference coarray compared to other sparse arrays, such as nested arrays and coprime arrays, etc.~\cite{liu2018optimizing}, which makes it less robust~\cite{essentialarray}. It is worth noting that the proposed approach is generic and can be applied to any sparse array configuration.
The sensor location vector $\mathbf{D}~\in\mathbb{R}^M$ is given by $\mathbf{D}=\{d_1,~d_2,~...,d_M\}$. Note that, the location of each sensor $d_i$ is chosen as an integer multiplied by a distance $d_0$, where $d_0$ is the smallest distance between any two adjacent sensors. As an example, for the MRA with four and five sensors, the sensors are located in the positions $\{0, 1, 4, 6\}d_0$ and $\{0, 2, 5, 8, 9\}d_0$, respectively. Further, we assume that there are $K$ uncorrelated narrow-band signals impinging on the array from directions $\Theta=[\theta_1,...,\theta_k,...\theta_K]^T$ with power $[\rho_1^2,...,\rho^2_k,...\rho^2_K]$. To this end, the steering vector for the $k-$th source is given by $\bm{a}(\theta_k)=[e^{jd_1\phi_k}~e^{jd_2\phi_k} \dots e^{jd_K\phi_k}].$ Here, $\phi_k=2\pi\text{sin}(\theta_k)/\lambda$ with $\lambda$ is the wavelength of the carrier frequency. Now, the received signal at time instant $t$, $\boldsymbol {y}(t)\in \mathbb{C}^M$ is given by 
\begin{equation}
\label{eq1}
    \boldsymbol {y}(t) = \boldsymbol {A}(\theta) \boldsymbol {x}(t) + \boldsymbol {n}(t), t = 1,2,\ldots,N, 
\end{equation}
 Here, $N$ is the total number of snapshots. Note that array steering matrix is given by $\boldsymbol {A}(\theta)=[\bm{a}(\theta)_1 \ldots \bm{a}(\theta)_K ]~\in \mathbb{C}^{M\times K}$. Here, $\boldsymbol {x}(t)=[x_1(t) \ldots x_K(t) ]^T$ and $\boldsymbol {n}(t)~\in \mathbb{C}^M$ are source signal vector and complex Gaussian noise, respectively. Note that we assume that source signals follow the unconditional model, i.e., the $\boldsymbol {x}(t)$ is a random signal~\cite{stoica1990performance}. With this assumption, the covariance matrix of the received signal is given by $\bar{\boldsymbol {R}}= \mathbb{E} [\boldsymbol {y}(t) \boldsymbol {y}^{H}(t)]~\in \mathbb{C}^{M\times M}$. Further, $\bar{\boldsymbol {R}}$ can be expressed as

 \begin{figure}
	\centering
	\includegraphics[width=1\linewidth]{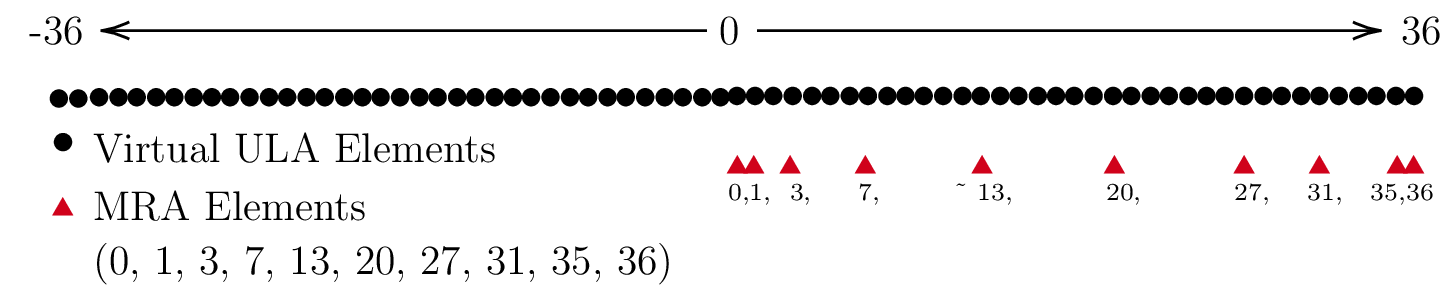}
	\caption{MRA of 10 physical sensors and corresponding $\mathcal{D}_{co}$  of the same aperture.}
	\label{fig:mra_fig}
\end{figure}

 \begin{equation} \bar{\boldsymbol {R}}=   \boldsymbol {A} \boldsymbol {R}_s \boldsymbol {A}^H + \sigma^2_n \boldsymbol {I}_M. \end{equation}
 Here,  the source covariance matrix is given by $\boldsymbol {R}_s= \mathbb{E} [\boldsymbol {x}(t) \boldsymbol {x}^{H}(t)]=\text{diag}\left(\rho_1^2,...\rho^2_K\right)$. Hence, the sample covariance matrix, which is the estimate of $\bar{\boldsymbol {R}}$ in practice, is defined as ${\bm{R}}=\frac{1}{N} \sum _{i=1}^{N}\bm{y}(t)\bm{y}(t)^H $. The coarray model can be then obtained through vectorizing ${\boldsymbol {R}}$, such that
  \begin{equation} \label{eq_rq}
  \boldsymbol {r}= \boldsymbol {A}_{co}\boldsymbol {\rho}+\sigma^2_n  \boldsymbol {e_i} ,
   \end{equation} 
   where $\boldsymbol {A}_{co}=\boldsymbol {A}^*\bigodot\boldsymbol {A}$ corresponds to the difference coarray steering vector, whose sensor locations are $\mathcal{D}_{co}=\{d_m-d_n|1\leq m,n\geq M\}$ , $\boldsymbol{\rho}=\left(\rho_1^2,...\rho^2_K\right)$ and $\boldsymbol {e_i}=\text{vec}(\boldsymbol {I}_M)$. Fig.~\ref{fig:mra_fig} shows an MRA of ten physical sensors and the corresponding difference coarray $\mathcal{D}_{co}$ of the same aperture. In this example, the constructed coarray can act as a virtual continuous ULA of 71 elements with no holes (i.e., hole free). The virtual sensor locations are defined by $[-M_v+1,-M_v+2,\dots,0,\dots,M_v-2,M_v-1]d_0$, where the size of the array is equal to $2M_v-1$. In Fig.~\ref{fig:mra_fig}, $M_v=36$.
   \subsection{Spatially-smoothed MUSIC (SS-MUSIC)} 
  Generally, in sparse arrays, $\mathbf{r}$ in eq.~\eqref{eq_rq} is considered as the received signal by the virtual ULA configuration previously mentioned, with corresponding steering matrix $\boldsymbol{A}_{co}$, and corresponding coarray covariance matrix $\mathbf{R}_{co}=\mathbf{r}\mathbf{r}^H$. However, in this case, $\mathbf{R}_{co}$ would be of rank one since the virtual source signal would become a single snapshot of  $\boldsymbol{A}_{co}$, similar to handling fully coherent sources \cite{Moeness}.  Therefore, conventional approaches such as MUltiple SIgnal Classification (MUSIC) cannot be directly applied to DOA estimation. Thus, spatial smoothing techniques would be required to restore the rank of the matrix \cite{wang2016coarrays} before applying MUSIC. To this end, the augmented covariance matrix after applying spatial smoothing explained in ~\cite{wang2016coarrays} is given by
  \begin{equation}
  \label{ss}
\boldsymbol {R}_\mathrm{ss}= \frac{1}{M_\mathrm{v}} \sum _{i=1}^{M_\mathrm{v}} \boldsymbol {r}_i \boldsymbol {r}_i^H,
  \end{equation}
where $\boldsymbol {R}_{ss}~\in \mathbb{C}^{M_v\times M_v}$. However, SS MUSIC is not resilient against elements failure, which will be explained in the next subsection.
   
 \begin{figure}
	\centering
	\includegraphics[width=1\linewidth]{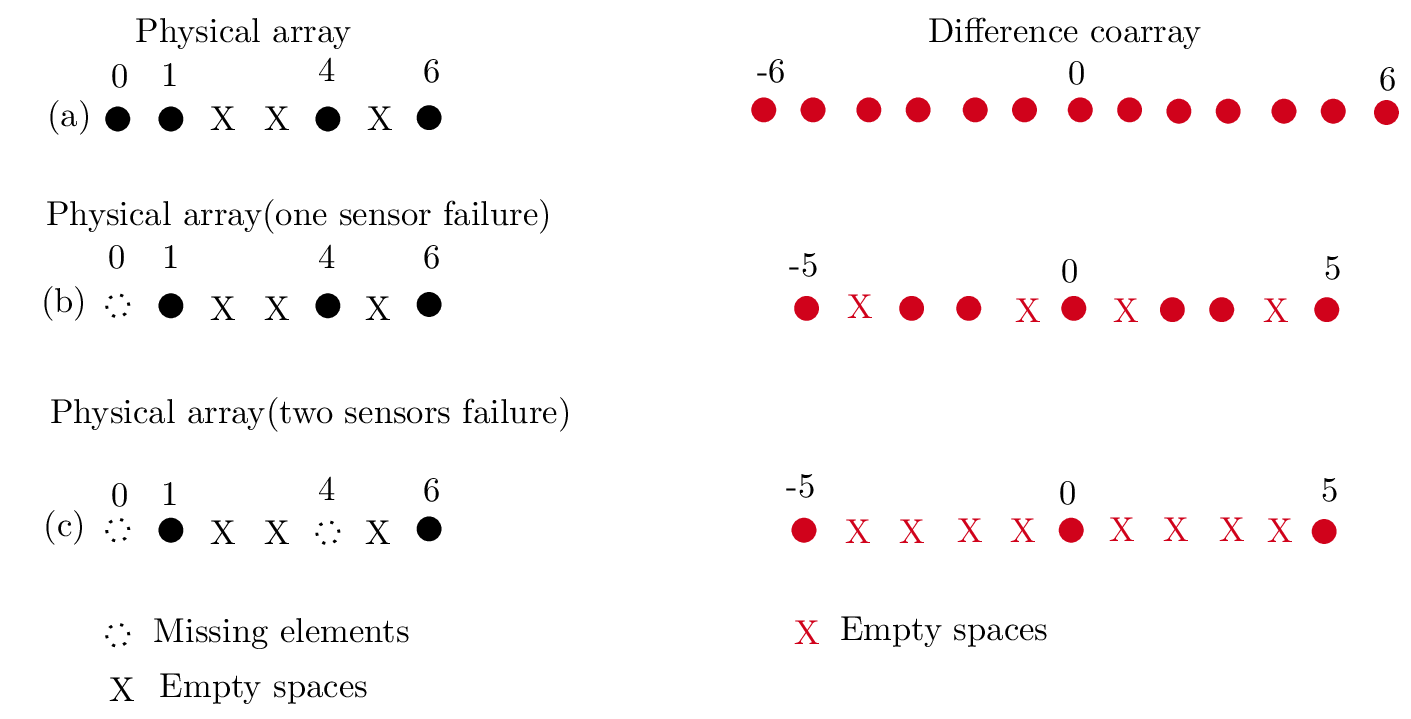}
	\caption{Impact of sensor failure of MRA.}
	\label{fig:MRA_MIS}
\end{figure}
\subsection{Effect of failures on $\mathcal{D}_{co}$}
  We now further investigate the effect of missing elements in the sparse array. For instance, in Fig~\ref{fig:MRA_MIS}, we consider an MRA with four sensors, and their locations are given by  \{$0$, $1$, $4$, $6$\}. 
This MRA and corresponding difference co-array are shown in Fig~\ref{fig:MRA_MIS}~(a). Now assume that the sensor in location $0$ failed, then due to this sensor failure, the difference co-array will not be hole-free. This is visualized in Fig~\ref{fig:MRA_MIS}~(b). Thus, as per \textit{Definition 2} in \cite{essentialarray}, this sensor is considered an essential element. If there is more than one essential sensor failure, then, the resultant co-array would have many holes~(Fig~\ref{fig:MRA_MIS}~(c)). Hence, if there is $M_1$ failed elements in the array, the corresponding received signal is modeled as if there is $M_1$ zero entries and similarly, the physical covariance matrix of the received signal ${\mathbf{R}} $ will have~$2MM_1-(M_1)^2$ missing elements in the matrix, modeled with zeros as well. We denote the covariance matrix of the received signal with sensor failures as $\boldsymbol {R}_m$, and the corresponding coarray to be $\mathbf{r}_m$. Furthermore, the augmented covariance matrix with missing elements after SS is denoted by $\mathbf{R}_{sm}$, \\  To tackle this problem, we propose two approaches. The first is hybrid, where we enhance the conventional SS MUSIC using deep neural network (DNN), to make it more resilient to sensor failures. The second approach is an end-to-end data-driven approach, where a DNN is used to estimate the missing elements of the  $\boldsymbol {R}_{m}$.
\section{the proposed DNN-based approaches}
\subsection{Hybrid (SS$\rightarrow$DNN)}

Hybrid approaches rely on proven signal processing methods and combine it with machine learning. 
Here, we improve the output of SS using a DNN. In more details, the algorithm consists of two blocks, the first one, applies the SS algorithm on the coarray received signal $\boldsymbol{r_m}$, where the values of the missing holes in the array are averaged. The second block, trains a feed forward DNN to further improve the covariance matrix. The proposed  DNN will take the $\mathbf{R}_{sm}$ matrix as an input and predict the SS augmented covariance matrix of a complete array $\mathbf{R}_{ss}$.\\
To train the DNN, we generate $P$ training samples of covariance matrices. 
Next, we vectorize the covariance matrix to convert it into a column vector. Then, the set of these column vectors is used to generate our data matrix, such that $\bar{\bm{Y}}_{sm}=[\text{vec}(\bm{R}_{sm}^1),\cdots,\text{vec}(\bm{R}_{sm}^P)]~\in ~\mathbb{C}^{M_v^2\times P}$ and $\bar{\bm{Y}}_{ss}=[\text{vec}(\bm{R}_{ss}^1),\cdots,\text{vec}(\bm{R}_{ss}^P)]~\in ~\mathbb{C}^{M_v^2\times P}$, where ${(.)}^i$ denotes the $i-$th training sample. \\ Furthermore, the real and imaginary components are cascaded together since the DNN is not designed to handle complex numbers, such that $\mathbf{Y}_{sm} = \left[\Re ({\bar{\mathbf{Y}}_{sm}})~\Im (\bar{{\mathbf{Y}}}_{sm})\right]~\in ~\mathbb{R}^{2M_v^2\times P}$ and $\mathbf{Y}_{ss} = \left[\Re ({\bar{\mathbf{Y}}_{ss}})~\Im (\bar{{\mathbf{Y}}}_{ss})\right] ~\in ~\mathbb{R}^{2M_v^2\times P}$. As a pre-processing step, data is normalized using min-max normalization. To simplify the notation, let  $2M_v^2=H$ and $2M^2=L$. The DNN is composed of four fully connected layers (i.e., an input layer, two hidden layers, and an output layer), each of size $H$. The number of hidden layers was chosen based on several trials. Further, to avoid over-fitting, a regularization technique is used. Specifically, dropout layers are added where they randomly drop the contribution of a certain percentage of neurons. In this structure, two dropouts were used with probabilities of $0.2$, $0.4$ after the input and first hidden layers respectively. The ReLU function is used as an activation function. Here, we train our DNN for $150$ epochs using the Adam optimizer. Furthermore, we consider the standard mean squared error as a loss function.
\subsection{Data Driven (only DNN)}


In this section, we propose an end-to-end data-driven algorithm, unlike the previous approach. Here, we propose a DNN only to compensate for the missing elements. The DNN is trained to compensate for the missing samples of the failed elements in the physical covariance matrix $\mathbf{R}_m$ and the same DNN is used to emulate the spatial smoothing process. Similar to the previous approach, the proposed DNN is composed of five fully connected layers, with a dropout layer of probabilities of $0.2$ after each layer. The DNN is trained with an input of the vectorized covariance matrix with missing data ($\bm{Y}_{m}$), and its output is covariance matrices with spatial smoothing~($\mathbf{Y}_{ss}$). Similar to the previous subsection, the real and imaginary are cascaded such that $\mathbf{Y}_{m} = \left[\Re (\bar{\mathbf{Y}}_{m})~\Im (\bar{\mathbf{Y}}_{m})\right]~\in ~\mathbb{R}^{2M^2\times P}$ where, $\bar{\bm{Y}}_{m}=[\text{vec}(\bm{R}_{m}^1),\cdots,\text{vec}(\bm{R}_{m}^P)]~\in ~\mathbb{C}^{M^2\times P}$.
Dimensions of the input layer and first hidden layer are $L$ and dimensions of the second, third hidden layers and output layer are $H$.

\subsection{DOA using MUSIC}
\begin{figure}
	\centering
	\includegraphics[width=1\linewidth]{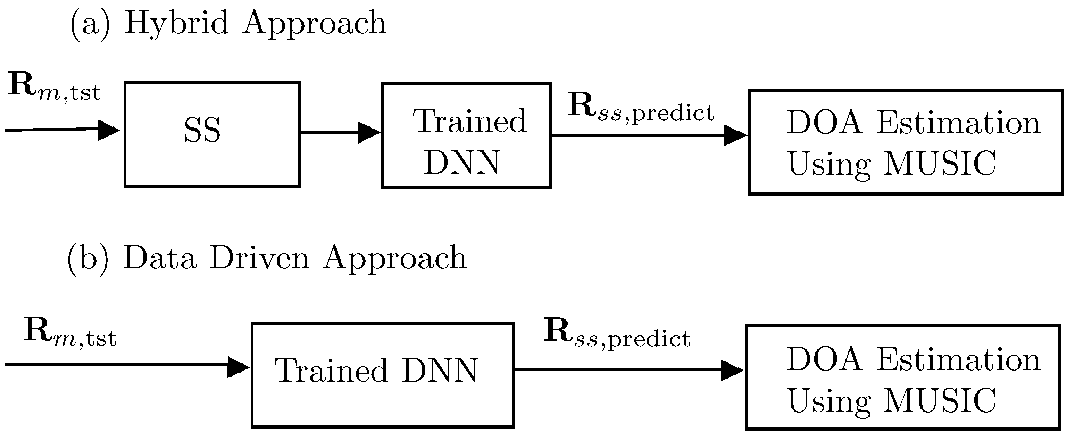}
	\caption{Proposed approaches for coarray matrix reconstruction in case of sensor failures.}
	\label{fig:dnn_rec}
\end{figure}

In both approaches, the DNN outputs predicted covariance matrices, which afterwards are fed to the conventional MUSIC algorithm for DOA estimation. Please note that there are end-to-end approaches to estimate DOA using machine learning, but this is out of the scope of this paper, since we mostly focus on compensating for the missing elements. To evaluate the performance of the DOA estimation, the average mean squared error (MSE) is used. It is given by 
\begin{equation}
\label{MSE}
    \text{MSE}= \frac{1}{KQ}\sum_{q=1}^{Q}\sum_{k=1}^{K} (\hat{\theta}_{q,k}-\theta_{q,k})^2. 
\end{equation}
Here, $Q$ is the number of testing trials. Further,   $\hat{\theta}_{q,k}$ and $\theta_{q,k}$  are the estimated and true angles of the $q-$th trial, respectively.
The testing process of both approaches is shown in Fig.~\ref{fig:dnn_rec}~(a) and (b), where the input of both approaches testing covariance matrix with missing elements $\mathbf{R}_{m,\text{tst}}$, and the output is predicted covariance matrix  ${\mathbf{R}}_{ss,\text{predict}}$. Next, this predicted covariance matrix are fed to the conventional MUSIC algorithm for DOA estimation. 

\begin{figure*}[!t]
	\centering
	\includegraphics[width=0.85\linewidth]{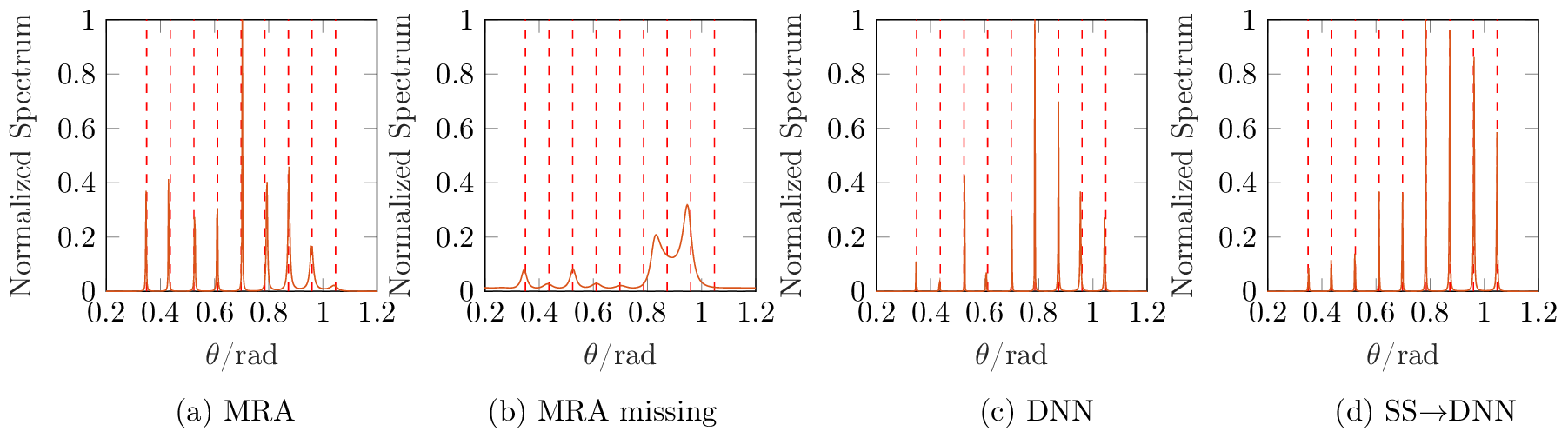}
	\caption{ MUSIC Spectrum for nine targets for SNR=$-10$ dB.}
	\label{fig:music_spec}
\end{figure*}
\section{Results}

\begin{figure}[!t]
	\centering
\includegraphics[width=1\linewidth]{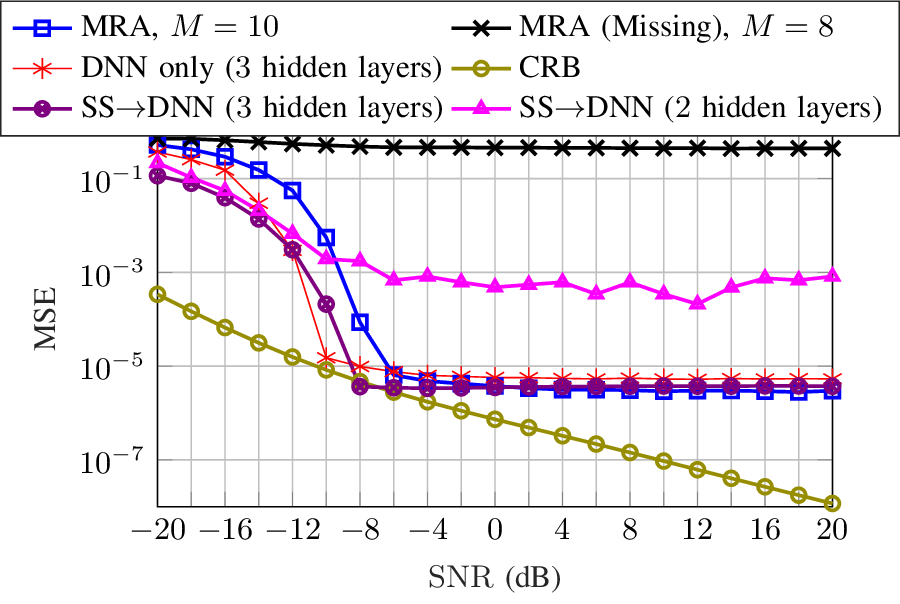}
\caption{DOA estimation comparison of the proposed approaches for\\~ $M1=2,~n=1,~m=4,~\text{and}~K=9$.} \label{fig_M_3}
\end{figure}

In the training phase, $p=3\times10^5$ training samples are generated. The data set is split into training and validation sets with ratios of $80\%$ and $20\%$, respectively. The number of physical antennas is $M=10$ and we consider nine targets ($K=9$) that can be randomly placed in the angle range interval $[10^{\circ},70^{\circ}]$. In the training phase, we set the angle gap between every two adjacent targets as $5$ degrees to avoid closely spaced targets since those cannot be separated using MUSIC.  Furthermore, we consider different signal-to-noise ratios (SNR) in the training data set. Here, SNR is selected uniformly on the interval of $[-10,10]$~dB. In addition, we assume that up to two sensors can fail, whose locations are unknown and denoted by $[d_n,d_m]$, where $1\leq n,m \leq M$. In the testing phase, our approach is evaluated over the SNR range of $[-20,20]$~dB with a step size $2$ dB. Here, in each SNR, we consider $Q=1000$ testing samples of covarianace matrices each with snapshots $N=200$. In the testing phase, we assume that the failure of the sensors occurs at $n=1$ and $m=5$. \\ First, we evaluate the MUSIC pseudospectrum of each of the proposed approaches compared to full MRA in Fig.~\ref{fig:music_spec}. Here, we set the SNR as $-10$~dB. Fig.~\ref{fig:music_spec}~(a) and Fig.~\ref{fig:music_spec}~(b) show the pseudospectrum using only SS MUSIC for both full MRA and MRA with two essential failed sensors. It can be shown that, sensor failures severely affect the DOA estimation in (Fig.~\ref{fig:music_spec}~(b)) and with sensor failure, the coarray has several holes in it, losing its high-resolution capability. However, the proposed hybrid and data-driven approaches were able to achieve similar performance as a full MRA with only $8$ physical sensors. One interesting note here is that both approaches were able to resolve one target more than the MRA. The MRA spectrum could not resolve all $9$ targets due to the low SNR. However, since the DNN in both proposed approaches were trained on many randomly chosen SNRs in the range $[-10,10]$ dB, it gained a denoising capability from the high SNRs. In more detail, the DNN learned an additional task to enhance the SNR of the predicted covariance matrix since it was trained on a various values of SNRs.  \\ In the following set of simulations, we further explore this behavior by conducting $1000$ Monte Carlo simulations. \\Fig.~\ref{fig_M_3} shows the average MSE of DOA estimation based on~\eqref{MSE} for MRA, MRA with missing sensors and the proposed approaches. Based on this figure, it can be observed that sensor failures severely affected the DOA estimation. However, the proposed approaches were able to improve DOA estimation in the presence of missing data. More specially, in the low SNR regime $\left([-20,-8]\right)$~dB, both approaches were able to outperform the MRA due to the denoising capabilities of the DNN. We observed this kind of denoising behavior also in our past study~\cite{ahmed2021deep}. \\
In addition, we have compared the hybrid approach using two and three hidden layers, to examine how SS can be beneficial to reduce the learning complexity compared to a data-driven approach. It can be noticed that at low SNRs, two hidden layers are enough to enhance the SS technique and recover the missing sensors. In more detail, it shows good behavior similar to the data-driven approach with three hidden layers. However, as the SNR increases, the data-driven approach along with the hybrid approach with three hidden layers have similar behavior, where they approach the MRA curve. It can be depicted that the hybrid approach with two hidden layers can provide good 
performance, when computational complexity matters the most, compared to the data-driven approach. This is useful when the system is running on low-end devices or has very limited resources. Furthermore, if we have prior knowledge that the system operates in the low SNR regime $[-20,-8]$~dB, we can utilize the hybrid approach which has similar performance as the data-driven approach. However, the latter approach is better when DOA accuracy matters the most since it provides the best performance compared to the original MRA array.  
\\ Further, we have compared the proposed methods with the Cram\'{e}r–Rao bound (CRB) for unbiased DOA estimation of MRA, which is given in~\cite{wang2016coarrays}
\begin{equation}
    \text{CRB}_{\theta}=\dfrac{1}{N}\left( \bm{M}^H_{\theta} (\bm{I}-\bm{M}_{S}(\bm{M}_{S}^H\bm{M}_{S})^{-1}\bm{M}^H_{S}) \bm{M}_{\theta}\right)^{-1}
\end{equation}
Where, $\bm{M}_{S}$ and  $\bm{M}_{\theta}$ are given by $(\bm{R}^T\otimes\bm{R})^{-1/2}\bm{\dot{A}}_d\bm{R}_s$ and $(\bm{R}^T\otimes\bm{R})^{-1/2}[\bm{A}_d~\text{vec}(\bm{I}_M)]$, respectively. Here, $\bm{\dot{A}}_d$ is given by $\bm{\dot{A}}^*\bigodot\bm{A}+\bm{A}^*\bigodot\bm{\dot{A}}$, where $\bm{\dot{A}}$ is the first derivative of the steering vector matrix $\bm{A}(\theta)$, given by 
$\bf{ \dot{A}}(\theta)=\left[\frac{\partial {\bf v}(\theta_{1})}{\partial\theta_{1}}, \frac{\partial {\bf v}(\theta_{2})}{\partial\theta_{2}},\cdots,\frac{\partial {\bf v}(\theta_{K})}{\partial\theta_{K}}\right]$.
It can be noticed that the actual MRA in high SNR saturates with higher MSE compared to the CRB, this is due to the fact that CRB depends on the SNR, where it may approach zero as SNR grows asymptotically large \cite{wang2016coarrays}. However, in reality, SS-MUSIC may have poor statistical properties when $K<M$, especially at high SNR.

\section{Conclusion}

Resilience is a critical KPI in many 6G applications. In this work, we address the resilience in MRAs with the aid of a DNN. Here, we proposed novel hybrid and data-driven based approaches to mitigate the impact of the sensors' failure in sparse arrays. The impact of these failures was investigated on the DOA estimation problem. In the results, we show that the proposed DNN-based approaches can achieve similar DOA estimation accuracy to the original array even in the presence of multiple failed sensors. Moreover, the proposed approaches show better performance in DOA estimation than the original sparse array with no failed elements in the low SNR regimes, thanks to the denoising capabilities of the DNN.


	\bibliographystyle{IEEEtran}\bibliography{IEEEabrv,ref.bib}

\begin{thebibliography}{10}
\providecommand{\url}[1]{#1}
\csname url@samestyle\endcsname
\providecommand{\newblock}{\relax}
\providecommand{\bibinfo}[2]{#2}
\providecommand{\BIBentrySTDinterwordspacing}{\spaceskip=0pt\relax}
\providecommand{\BIBentryALTinterwordstretchfactor}{4}
\providecommand{\BIBentryALTinterwordspacing}{\spaceskip=\fontdimen2\font plus
\BIBentryALTinterwordstretchfactor\fontdimen3\font minus
  \fontdimen4\font\relax}
\providecommand{\BIBforeignlanguage}[2]{{%
\expandafter\ifx\csname l@#1\endcsname\relax
\typeout{** WARNING: IEEEtran.bst: No hyphenation pattern has been}%
\typeout{** loaded for the language `#1'. Using the pattern for}%
\typeout{** the default language instead.}%
\else
\language=\csname l@#1\endcsname
\fi
#2}}
\providecommand{\BIBdecl}{\relax}
\BIBdecl

\bibitem{6G_JCS}
G.~Fettweis \emph{et~al.}, ``{WHITE PAPER: Joint Communications \& Sensing},''
  Tech. Rep., 07 2021.

\bibitem{6G}
K.Trommler \emph{et~al.}, ``{WHITE PAPER: Six Insights into {6G} Orientation
  and Input for Developing Your Strategic {6G} Research Plan},''
  {\"U}bernationale Vereinigung f{\"u}r Kommunikationsforschung e.V, Tech.
  Rep., 05 2022.

\bibitem{robertspaper}
\BIBentryALTinterwordspacing
R.-J. Reifert, S.~Roth, A.~A. Ahmad, and A.~Sezgin, ``Comeback {Kid}:
  Resilience for mixed-critical wireless network resource management,'' 2022.
  [Online]. Available: \url{https://arxiv.org/abs/2204.11878}
\BIBentrySTDinterwordspacing

\bibitem{sparsearray}
S.~Qin, Y.~D. Zhang, and M.~G. Amin, ``Generalized coprime array configurations
  for direction-of-arrival estimation,'' \emph{IEEE Trans. Signal Proces.},
  vol.~63, no.~6, pp. 1377--1390, 2015.

\bibitem{MRA}
A.~Moffet, ``Minimum-redundancy linear arrays,'' \emph{IEEE Trans. Antennas
  Propag.}, vol.~16, no.~2, pp. 172--175, 1968.

\bibitem{nestedarrays}
P.~Pal and P.~P. Vaidyanathan, ``{Nested Arrays: A Novel Approach to Array
  Processing With Enhanced Degrees of Freedom},'' \emph{IEEE Trans. Signal
  Proces.}, vol.~58, no.~8, pp. 4167--4181, 2010.

\bibitem{MRAoptimization}
C.-L. Liu and P.~P. Vaidyanathan, ``Optimizing minimum redundancy arrays for
  robustness,'' in \emph{2018 52nd Asilomar Conf. Signals, Syst., Comput.},
  2018, pp. 79--83.

\bibitem{essentialarray}
------, ``Comparison of sparse arrays from viewpoint of coarray stability and
  robustness,'' in \emph{2018 IEEE 10th Sensor Array and Multichannel Signal
  Processing Workshop (SAM)}, 2018, pp. 36--40.

\bibitem{Stoica_missingdata}
E.~Larsson and P.~Stoica, ``High-resolution direction finding: the missing data
  case,'' \emph{IEEE Trans. Signal Proces.}, vol.~49, no.~5, pp. 950--958,
  2001.

\bibitem{SOA_1}
B.~Sun, C.~Wu, J.~Shi, H.-L. Ruan, and W.-Q. Ye, ``Direction-of-arrival
  estimation under array sensor failures with {ULA},'' \emph{IEEE Access},
  vol.~8, pp. 26\,445--26\,456, 2020.

\bibitem{missingelement}
M.~Wang, Z.~Zhang, and A.~Nehorai, ``Direction finding using sparse linear
  arrays with missing data,'' in \emph{2017 IEEE Int. Conf. Acoust., Speech,
  Signal Process. (ICASSP)}, 2017, pp. 3066--3070.

\bibitem{liu2018optimizing}
C.-L. Liu and P.~Vaidyanathan, ``Optimizing minimum redundancy arrays for
  robustness,'' in \emph{2018 52nd Asilomar Conf. Signals, Syst.,
  Comput.}\hskip 1em plus 0.5em minus 0.4em\relax IEEE, 2018, pp. 79--83.

\bibitem{stoica1990performance}
P.~Stoica and A.~Nehorai, ``Performance study of conditional and unconditional
  direction-of-arrival estimation,'' \emph{IEEE Trans. on Acoustics, Speech,
  and Signal Processing}, vol.~38, no.~10, pp. 1783--1795, 1990.

\bibitem{Moeness}
S.~Qin, Y.~D. Zhang, and M.~G. Amin, ``Generalized coprime array configurations
  for direction-of-arrival estimation,'' \emph{IEEE Trans. Signal Proces.},
  vol.~63, no.~6, pp. 1377--1390, 2015.

\bibitem{wang2016coarrays}
M.~Wang and A.~Nehorai, ``Coarrays, {MUSIC}, and the {Cram{\'e}r--Rao} bound,''
  \emph{IEEE Trans. Signal Process.}, vol.~65, no.~4, pp. 933--946, 2016.

\bibitem{ahmed2021deep}
A.~M. Ahmed, U.~S. K.~M. Thanthrige, A.~El~Gamal, and A.~Sezgin, ``Deep
  learning for {DOA} estimation in {MIMO} radar systems via emulation of large
  antenna arrays,'' \emph{IEEE Commun. Lett.}, vol.~25, no.~5, pp. 1559--1563,
  2021.

\end{thebibliography}

\end{document}